\documentclass[sigconf,nonacm]{acmart}
\usepackage{booktabs}
\usepackage{multirow}
\usepackage{tikz}
\usetikzlibrary{arrows.meta,positioning}
\usepackage{pgfplots}
\pgfplotsset{compat=1.18}
\definecolor{cpolicy}{HTML}{2a78d6}
\definecolor{ctool}{HTML}{eb6834}
\definecolor{cdata}{HTML}{1baf7a}
\definecolor{cmodel}{HTML}{eda100}
\definecolor{cink}{HTML}{52514e}

\begin{document}

\title{Can You Trust Your Finance Agent? Drift, Diagnosis and Auditable Intervention on One Shipped Financial Research Agent}

\author{Cosimo Galeone}
\authornotemark[1]
\affiliation{\institution{}\country{}}
\email{cosimo.galeone@alomana.com}
\author{Anna Ettorre}
\authornotemark[1]
\affiliation{\institution{}\country{}}
\email{anna.ettorre@alomana.com}
\author{Ginevra Mariani}
\authornotemark[1]
\affiliation{\institution{}\country{}}
\email{ginevra.mariani@alomana.com}
\author{Minsu Park}
\authornotemark[1]
\affiliation{\institution{}\country{}}
\email{minsu.park@alomana.com}
\author{Giuseppe Ettorre}
\authornotemark[1]
\affiliation{\institution{}\country{}}
\email{ge@alomana.com}
\author{Daniele Ligorio}
\authornotemark[1]
\affiliation{\institution{}\country{}}
\email{dl@alomana.com}

\begin{teaserfigure}
\makebox[\textwidth][c]{\normalsize\textsuperscript{*}Alomana, Grottaglie, Italy}
\end{teaserfigure}

\begin{abstract}
Can you trust a finance agent you did not build? We take the reference agent shipped with the Vals Finance Agent Benchmark v2, leave it untuned, and ask four questions a leaderboard score does not answer. Is the score stable? Seven identical runs on 18 questions pass between 105 and 123 of 149 rubric criteria, and the bill varies by up to 17\%. When the agent fails, who owns the fix? We classify 20 failed trajectories, from three runs of nine held-out questions, by the layer whose owner would remedy them, under a taxonomy frozen in advance: 13 belong to the agent's instructions, 3 to its tools, 4 to the model, none to its data. Can that diagnosis be trusted? Four blind readers from three model families, given only the traces, name the same fault as our LLM analyzer in 94\% of cases, the rate at which they agree with one another (93\%). Can a change be trusted? An optimization loop reads the traces, makes one targeted change per round and keeps it only if it clears a fixed threshold. Under a cost objective its final configuration costs 0.34$\times$ the baseline on the held-out questions (95\% CI [0.22, 0.58]) at accuracy within noise. Under an accuracy objective it gains 14 points on the development questions and 1.5 on the held-out ones (CI [$-$5.3, +7.2]). Judged against the measured run-to-run noise, under a rule written before the test was opened, the first is a real saving and the second is not. This is a case study of one agent and nine test questions, not a general result.
\end{abstract}


\maketitle

\section{Introduction}
\label{sec:intro}

A financial research agent is in production. It has a leaderboard score. A compliance officer asks what it does when it is wrong. The score does not answer: it was computed once, on questions the officer has never seen, by a grader the institution does not control, and it says nothing about where the wrong answers came from or who in the building should fix them.

Benchmarks of financial agents have multiplied since 2025. The Vals Finance Agent Benchmark~\cite{bigeard2025fab} scores answers to expert-written research questions against rubric criteria; BigFinanceBench~\cite{wang2026bigfinancebench} grades the visible derivation step by step. Both measure how much of the answer, or of its derivation, is right. Neither says why a wrong answer was wrong, and neither names the layer of the system that owns the remedy. DFAH-Bench~\cite{khatchadourian2026dfah} comes closest to the question we ask: replaying the same financial decision task, it finds that decisions agree 94--95\% of the time while exact tool paths agree only 67--69\%. The processes diverge; the benchmark establishes it and stops there.

We stop later. This paper is a controlled case study on one shipped agent that we did not write and did not tune, organised as five questions, each answering what the previous one leaves open.

\begin{enumerate}
  \item \emph{Is the score a property of the agent?} No. Seven identical runs of the shipped configuration span 105--123 of 149 criteria (\S\ref{sec:drift}).
  \item \emph{When it fails, who owns the fix?} On a pre-registered set of 20 failed trajectories, under our owner-of-the-remedy taxonomy: policy 13, tool 3, data 0, model 4 (\S\ref{sec:anatomy}).
  \item \emph{Can you trust that diagnosis?} Blind readers of three model families name the same fault in 75 of 80 cases; the layer is a definition we supply; localisation is looser (\S\ref{sec:panel}).
  \item \emph{Can you trust a change?} Same loop, two objectives: cost falls to 0.34$\times$ and holds on the frozen test; accuracy rises 14 points on dev and is indistinguishable from zero on the test (\S\ref{sec:loop}).
  \item \emph{What is trust, then?} A protocol whose every step can be audited (\S\ref{sec:trail}).
\end{enumerate}

The finance cut on the second question is the owner of the remedy. In an institution, compliance rewrites the agent's policy, the platform engineer owns the tool, the data owner the retrieval pipeline, and model risk the model. We classify failures by that owner, not by symptom, and we measure rather than assert what the classification means (\S\ref{sec:panel}).

\paragraph{Scope.} This is a case study of one agent: nine test questions and twenty failed trajectories give an anatomy rather than population shares, and every test-split interval is a paired bootstrap over the nine questions. The fourth question is where the first three are put to work. Cost reduction is not the contribution of this paper; it is the intervention the diagnosis pointed at, on the axis where our instruments can resolve an effect (0.34 against a run-to-run noise band of 0.83). On accuracy the same instruments return noise, and that is a result too.

We contribute (i) a drift measurement of a shipped agent across seven identical baselines; (ii) an anatomy of its failures on a pre-registered set, under a taxonomy frozen before the test split was analysed; (iii) a blind, cross-family check of the diagnostic that separates fault identification (corroborated), remedy ownership (a definition, reproducible from the rule) and localisation (looser), with the diagnostic's own errors declared; (iv) the optimization loop as shipped, under two objectives, each scored once on a frozen split under a written rule; (v) the protocol itself. We do not claim a better agent, a general method, or comparability with any leaderboard (\S\ref{sec:limits}).

\begin{figure}[t]
\centering
\begin{tikzpicture}[
  node distance=5mm and 4mm,
  box/.style={draw=cink, rounded corners=1pt, align=center, font=\footnotesize, inner sep=3pt, minimum height=9mm, text width=21mm},
  frz/.style={box, fill=cpolicy!15, text width=33mm},
  lab/.style={font=\scriptsize, color=cink, align=center},
  arr/.style={-{Stealth[length=1.6mm]}, cink, thick}]
\node[box] (agent) {agent\\(shipped config.)};
\node[box, right=of agent] (judge) {judge\\scores answers};
\node[box, right=of judge] (analyzer) {analyzer\\reads traces};
\node[box, below=7mm of judge] (loop) {loop: one change,\\gate keeps it or not};
\draw[arr] (agent) -- (judge);
\draw[arr] (judge) -- (analyzer);
\draw[arr] (analyzer) |- node[lab, pos=0.25, right] {diagnosis} (loop);
\draw[arr] (loop) -| node[lab, pos=0.75, left] {new\\config.} (agent);
\node[lab, above=1mm of judge] {optimization cycle, 18 dev questions only};
\node[frz, anchor=north] (test) at ([xshift=-19mm, yshift=-8mm]loop.south) {frozen test split, 9 questions:\\final config.\ vs baseline,\\rule written in advance};
\node[frz, anchor=north] (tax) at ([xshift=19mm, yshift=-8mm]loop.south) {frozen taxonomy:\\who owns the fix\\(\S\ref{sec:anatomy})};
\draw[arr] (loop) -- node[lab, left, pos=0.4] {final config.} (test);
\draw[arr] (analyzer.south) ++(4mm,0) -- node[lab, right, pos=0.5] {test-baseline\\diagnoses} ([xshift=4mm]analyzer.south |- tax.north);
\end{tikzpicture}
\caption{The system and the protocol. Agent, judge, analyzer and loop form the optimization cycle of \S\ref{sec:loop}, run on the dev split only. Shaded boxes are frozen before use: the test split and its decision rule before any test number existed (\S\ref{sec:test}), the taxonomy before the test trajectories were analysed (\S\ref{sec:taxonomy}).}
\label{fig:protocol}
\end{figure}
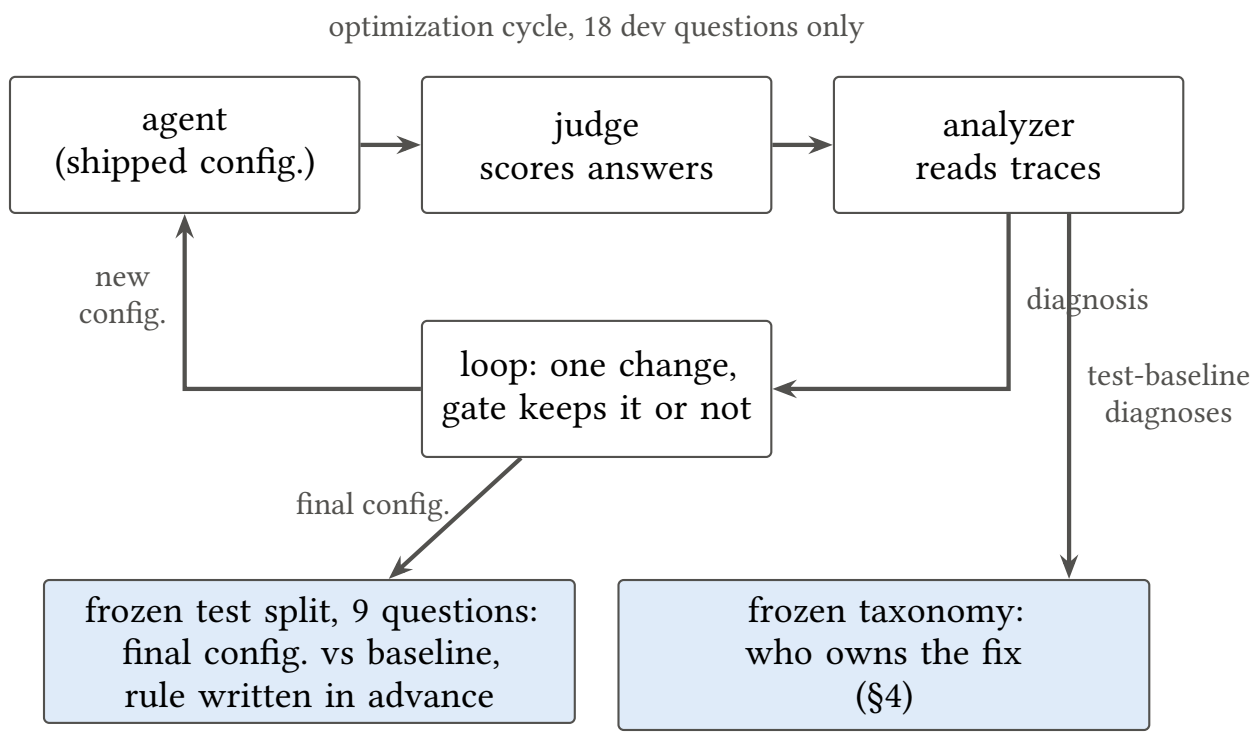

\section{Setting}
\label{sec:setting}

\paragraph{Terms.} The \emph{agent} is the shipped configuration under evaluation. The \emph{analyzer} is the component that reads a scored run's traces and emits a diagnosis. The \emph{loop} is the component that proposes one change per round from that diagnosis, applies it and accepts or rejects it. The \emph{protocol} is the set of freezes, replicas, rules and test openings in Fig.~\ref{fig:protocol}. The \emph{audit trail} is the set of files the protocol produced.

\paragraph{Models.} For reference, the models in this paper and their roles: the agent runs on Claude Sonnet 4.5~\cite{anthropic2025sonnet45}, including its nested retrieval call, as shipped, and the cost arm of \S\ref{sec:cost} moves that call to Claude Haiku 4.5~\cite{anthropic2025haiku45}; the judge is Gemini 3.8 Flash~\cite{google2026gemini38flash}; the analyzer, and the loop's proposer and applier, are Claude Sonnet 5~\cite{anthropic2026sonnet5}; the blind readers of \S\ref{sec:blind} are Claude Opus 5~\cite{anthropic2026opus5}, Gemini 3.1 Pro (preview)~\cite{google2026gemini31pro}, Gemini 3.8 Flash and GPT-5.5~\cite{openai2026gpt55}, and the model that compares their justifications is Claude Opus 5. One reader, Gemini 3.8 Flash, is the same model as the judge.

\paragraph{The agent.} The Vals Finance Agent Benchmark v2 ships a runnable agent: a tool-calling loop over Claude Sonnet 4.5 with seven tools (web search, EDGAR full-text search, HTML page parsing, a retrieval tool that reads stored filings through a nested LLM call, a calculator, price history, and final-answer submission) and a 2{,}767-character system prompt~\cite{bigeard2025fab}. Its public split has 27 questions across nine categories, each with rubric criteria carrying a severity and a must-pass flag. We reproduce the shipped configuration with one documented backend substitution, required for reproducibility: the agent's EDGAR search depends on a paid third-party index, which we replaced with the SEC's own full-text search endpoint behind the same tool name, description and argument schema. On the 53 EDGAR queries of one baseline run replayed on both backends, the target company's main filings were returned by both in 81 of 81 cases; the comparison is in Appendix~\ref{app:edgar}. Every run exports its trajectory in the Agent Trajectory Interchange Format~\cite{harbor2026atif}, which is what the analyzer reads.

\paragraph{The split.} We froze an 18/9 dev/test split, stratified across the nine categories and issuer-disjoint, before any analyzer ran. Every configuration change in this paper was decided on the 18 dev questions; the 9 test questions (Table~\ref{tab:test-questions}) were opened twice, each time under a protocol written before any test number was seen, and are now closed.

\begin{table}[t]
\caption{The nine test questions (Vals FABv2 public split).}
\label{tab:test-questions}
\small
\begin{tabular}{@{}lll@{}}
\toprule
idx & category & issuers \\
\midrule
1 & General qualitative analysis & MAR, WH \\
4 & General quantitative analysis & CZR \\
8 & Market analysis & HUBS, NOW, TOST \\
9 & Comparables & AER, FTAI, WLFC \\
14 & Precedents & PFE, SGEN \\
17 & Adjustments & ASPI \\
18 & Earnings analysis & VSEC \\
21 & Disclosure analysis & PFE \\
24 & Financial modeling & CRM \\
\bottomrule
\end{tabular}
\end{table}

\paragraph{The judge.} We score against the benchmark's verbatim criteria with our own judge: Gemini 3.8 Flash at temperature 0, three votes per criterion, a 0.5\% tolerance on numeric matches, reading the final answer only. Its configuration fingerprint is identical across every report in this paper. Two facts bound the judge's role. The cost results of \S\ref{sec:loop} are sums of provider prices per call: they do not depend on the judge's verdicts, so no judge error can move them. The pass-rate results do, and the judge is internally consistent under our fixed configuration: of the three votes cast per criterion, the vote split in 4 of 1{,}080 test cells and 12 of 1{,}043 dev baseline cells. This establishes within-configuration stability, not agreement with an external or human judge. Our numbers are internally comparable and not comparable with the Vals leaderboard, whose grader we have not used. We have not re-scored the test reports with a second judge model; the stored final answers allow it without re-running the agent, and \S\ref{sec:limits} says so.

\paragraph{The analyzer and the loop.} We use AloLab~\cite{galeone2026correct}, an evaluation and optimization engine, as a black box: we configure it and read its outputs, and we did not modify it. The analyzer (Claude Sonnet 5) reads a scored run and emits failure categories, each with a location in the trace (a decisive step), a \texttt{fix\_level} (prompt, skill, tool description, orchestration, architecture, or data issue) and a \texttt{harness\_component} (observation, context, control, action, state, verification, or none). The loop proposes one change per round on one of six declared levers (the system prompt, tool descriptions, which tools are enabled, the agent's model, the nested retrieval model, and that model's instructions), applies it, re-runs the dev split, and promotes or rejects the change against the round's parent configuration under a fixed gate. We wrote no agent and tuned nothing by hand.

\section{The score is a draw}
\label{sec:drift}

\begin{figure*}[t]
\centering
\begin{minipage}[t]{0.46\textwidth}\centering
\begin{tikzpicture}
\begin{axis}[width=\linewidth, height=3.4cm, scale only axis=false, ybar, bar width=4pt, ymin=95, ymax=135,
  xtick={1,...,7}, xlabel={\scriptsize baseline run}, ylabel={\scriptsize cells passed / 149},
  tick label style={font=\tiny}, label style={font=\scriptsize}, axis line style={cink}, ymajorgrids, grid style={cink!20},
  nodes near coords, every node near coord/.append style={font=\tiny, color=cink}]
\addplot[fill=cpolicy, draw=none] coordinates {(1,119) (2,116) (3,120) (4,123) (5,105) (6,116) (7,109)};
\end{axis}
\end{tikzpicture}\\[-2pt]{\scriptsize (a) Dev: drift over seven identical runs}
\end{minipage}\hfill
\begin{minipage}[t]{0.46\textwidth}\centering
\begin{tikzpicture}
\begin{axis}[width=\linewidth, height=3.4cm, ybar stacked, bar width=9pt, ymin=0, ymax=8,
  xtick={1,2,3}, xticklabels={r1,r2,r3}, xlabel={\scriptsize test baseline replica}, ylabel={\scriptsize failed trajectories},
  tick label style={font=\tiny}, label style={font=\scriptsize}, axis line style={cink}, ymajorgrids, grid style={cink!20},
  legend style={font=\tiny, draw=none, fill=none, at={(0.5,1.02)}, anchor=south, legend columns=3},
  nodes near coords, every node near coord/.append style={font=\tiny, color=white}, point meta=explicit]
\addplot[fill=cpolicy, draw=white, line width=0.6pt] coordinates {(1,3) [3] (2,6) [6] (3,4) [4]};
\addplot[fill=ctool, draw=white, line width=0.6pt] coordinates {(1,1) [1] (2,1) [1] (3,1) [1]};
\addplot[fill=cmodel, draw=white, line width=0.6pt] coordinates {(1,3) [3] (2,0) [] (3,1) [1]};
\legend{policy, tool, model}
\end{axis}
\end{tikzpicture}\\[-2pt]{\scriptsize (b) Test: who owns the fix, per baseline replica (data = 0)}
\end{minipage}

\vspace{3mm}
\begin{minipage}[t]{0.46\textwidth}\centering
\begin{tikzpicture}
\begin{axis}[width=\linewidth, height=3.4cm, xmin=0.4, xmax=2.6, ymin=-12, ymax=12,
  xtick={1,2}, xticklabels={cost, accuracy}, xlabel={\scriptsize winner}, ylabel={\scriptsize $\Delta$ pass rate vs baseline (pts)},
  tick label style={font=\tiny}, label style={font=\scriptsize}, axis line style={cink}, ymajorgrids, grid style={cink!20}]
\addplot[cink, dashed, domain=0.4:2.6] {0};
\addplot[only marks, mark=*, mark size=2pt, cpolicy, error bars/.cd, y dir=both, y explicit, error bar style={cpolicy, line width=1pt}]
  coordinates {(1,-3.7) += (0,4.8) -= (0,5.2) (2,1.5) += (0,5.7) -= (0,6.8)};
\end{axis}
\end{tikzpicture}\\[-2pt]{\scriptsize (c) Test: accuracy, 95\% CI over 9 questions}
\end{minipage}\hfill
\begin{minipage}[t]{0.46\textwidth}\centering
\begin{tikzpicture}
\begin{axis}[width=\linewidth, height=3.4cm, xmin=0.4, xmax=2.6, ymin=0.1, ymax=1.7,
  xtick={1,2}, xticklabels={cost, accuracy}, xlabel={\scriptsize winner}, ylabel={\scriptsize cost ratio vs baseline},
  tick label style={font=\tiny}, label style={font=\scriptsize}, axis line style={cink}, ymajorgrids, grid style={cink!20}]
\addplot[fill=cink!15, draw=none] coordinates {(0.4,0.83) (2.6,0.83) (2.6,1.205) (0.4,1.205)} \closedcycle;
\addplot[cink, dashed, domain=0.4:2.6] {1};
\addplot[only marks, mark=*, mark size=2pt, ctool, error bars/.cd, y dir=both, y explicit, error bar style={ctool, line width=1pt}]
  coordinates {(1,0.341) += (0,0.239) -= (0,0.125) (2,1.082) += (0,0.423) -= (0,0.238)};
\node[font=\tiny, cink] at (axis cs:1.5,1.0) {noise band};
\end{axis}
\end{tikzpicture}\\[-2pt]{\scriptsize (d) Test: cost, 95\% CI; band = [0.83, 1/0.83]}
\end{minipage}
\caption{The four measurements. Top left on the dev split, the other three on the frozen test split. (a) Cells passed by the shipped configuration in seven identical dev runs. (b) Primary layer of the 20 failed test trajectories per baseline replica. (c, d) Both loop winners on the frozen test, three replicas each, paired bootstrap over the nine questions: the cost winner's cost interval lies entirely below the drift band; every accuracy interval includes zero.}
\label{fig:main}
\end{figure*}
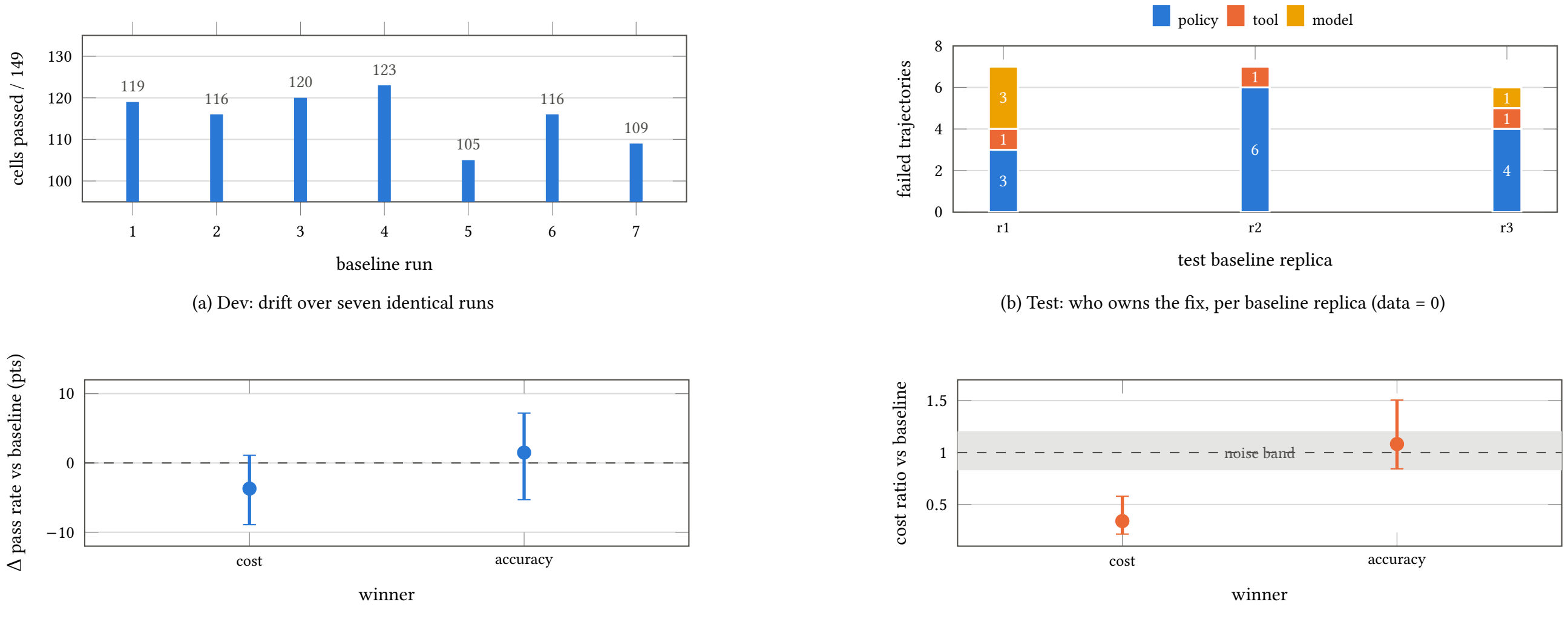

Before asking whether a change helped, one has to know how much the agent moves on its own. A score is, in this sense, a draw: one sample from the distribution over repeated runs of a fixed configuration, not a property of the configuration. We ran the shipped configuration seven times on the 18 dev questions and scored every run with the same judge (Table~\ref{tab:drift}, Fig.~\ref{fig:main}a).

\begin{table}[t]
\caption{Seven identical runs of the shipped configuration on 18 dev questions, 149 rubric criteria. Twins are consecutive baselines; a flip is a criterion passed in one run and not the other.}
\label{tab:drift}
\footnotesize
\begin{tabular}{@{}lrrr@{}}
\toprule
run & cells/149 & pass rate & agent \$ \\
\midrule
1 & 119 & 0.799 & 34.42 \\
2 & 116 & 0.779 & 28.58 \\
3 & 120 & 0.805 & 32.48 \\
4 & 123 & 0.826 & 32.28 \\
5 & 105 & 0.705 & 31.79 \\
6 & 116 & 0.779 & 32.98 \\
7 & 109 & 0.732 & 29.42 \\
\midrule
twin & flips & ratio of totals & geomean \\
\midrule
1$\to$2 & 19 & 0.830 & 0.867 \\
3$\to$4 & 21 & 0.994 & 0.876 \\
4$\to$5 & 28 & 0.985 & 1.044 \\
5$\to$6 & 15 & 1.038 & 0.924 \\
6$\to$7 & 19 & 0.892 & 1.004 \\
\bottomrule
\end{tabular}
\end{table}

Cells passed span 105 to 123 of 149: 18 criteria, twelve points of pass rate, with nothing changed. Between two consecutive runs 15 to 28 criteria flip. The bill moves by up to 17\% (ratio of run totals 0.830 to 1.038), and per question by far more: across the five twin pairs, a single question cost anywhere from 0.16$\times$ to 5.43$\times$ what it had cost the run before. Two consecutive runs, 6 and 7, scored 116 and 109.

The test split drifts the same way. Its three baseline replicas passed 54, 56 and 61 of 90 cells and cost \$20.58, \$23.67 and \$26.91.

Two consequences follow for anyone who promotes a change. The loop's default promotion bars sit inside this drift: a change is promoted if it cuts cost by at least 5\% under the cost objective, or raises the pass rate by at least one point under the accuracy objective, each measured on a single run against its parent. In \S\ref{sec:acc} a +7-cell change is promoted, and +7 is the distance between baselines 6 and 7. And a median or geometric-mean gate cannot see what happens per question: between runs 3 and 4 the total bill moved by less than 1\%, while per-question cost ratios ran from 0.16$\times$ to 2.52$\times$. A change that makes the twelve cheapest questions 30\% cheaper and the three dearest 60\% dearer passes a geometric-mean gate at 0.853 while the bill rises 11\%; we built that case on the measured per-question costs of baseline 3, and the gate would approve a change that did nothing.

One failure class in the drift has a compliance name. The agent parses the wrong fiscal period's filing: for one dev issuer (AMZN) in six of the seven baselines, for three others (DKNG, LULU, WYNN) depending on the draw. It is a property of the run, not of the configuration.

\section{Who owns the fix}
\label{sec:anatomy}

\subsection{Taxonomy}
\label{sec:taxonomy}

We classify each failed trajectory by the layer that owns its remedy: \textbf{policy} (the instructions the agent is given; owner, whoever writes its policy), \textbf{tool} (interfaces, descriptions, output parsing, state between steps; owner, the platform engineer), \textbf{data} (what reaches the model's context: sources, retrieval, truncation, staleness; owner, the data owner), and \textbf{model} (the model itself, given the right instructions, tools and facts; owner, model risk). The analyzer does not speak this vocabulary. It emits \texttt{fix\_level} and \texttt{harness\_component}, and we project the pair onto our four layers with a fixed table (Table~\ref{tab:projection}) built from the analyzer's own documentation and frozen before any analyzer run on the test trajectories.

\begin{table}[t]
\caption{The projection, frozen before the test trajectories were analysed. \texttt{fix\_level} is the primary key; \texttt{harness\_component} decides only where the key is ambiguous.}
\label{tab:projection}
\small
\begin{tabular}{@{}lll@{}}
\toprule
\texttt{fix\_level} & \texttt{harness\_component} & layer \\
\midrule
prompt & any but none & policy \\
prompt & none & policy or model, by Rule M \\
skill & any & policy \\
tool description & any & tool \\
orchestration & action, observation, state & tool \\
orchestration & context & data \\
orchestration & control, verification, none & policy \\
architecture & context & data \\
architecture & other, none & tool \\
data issue & any & benchmark artefact; outside \\
\bottomrule
\end{tabular}
\end{table}

One cell needs a rule. A \texttt{prompt} category with no harness component is the analyzer's ``pure prompt or reasoning'' finding, and it is exactly where the policy/model boundary lies. \emph{Rule M}: the category maps to model when both hold, read from the trace and not from the analyzer's prose: (a) the analyzer names no missing or wrong instruction, and (b) the decisive step is a reasoning, assistant or output event, and the facts it needed were already in the context at that step. Otherwise policy. Model is the residual: the case a model swap, and no harness change, would remedy.

The unit of count is the failed trajectory. One trajectory may carry several categories; its primary layer is the layer of the category containing its decisive step. A trajectory whose only category is a data issue is a benchmark artefact and leaves the denominator. Four choices in the table have defensible alternatives; we record them in Appendix~\ref{app:taxonomy}. Two \texttt{fix\_level} values the analyzer emitted on the dev split (\texttt{retrieval\_llm}, \texttt{tool\_descriptions}) are outside its documented vocabulary and are left unmapped; neither occurs on the test split.

\subsection{The count}
\label{sec:count}

The pre-registered set is the three baseline replicas of the test split: 27 trajectories, 20 of them failed. Table~\ref{tab:count} and Fig.~\ref{fig:main}b give the count; Appendix~\ref{app:tables} gives every trajectory's decisive step and verdict.

\begin{table}[t]
\caption{Primary layer per failed trajectory, three test baseline replicas. Reading A is Rule M as written; reading B additionally treats calculator arguments as model-authored and is a sensitivity only.}
\label{tab:count}
\small
\begin{tabular}{@{}lrrrrrr@{}}
\toprule
replica & failed & policy & tool & data & model (A) & model (B) \\
\midrule
r1 & 7 & 3 & 1 & 0 & 3 & 4 \\
r2 & 7 & 6 & 1 & 0 & 0 & 0 \\
r3 & 6 & 4 & 1 & 0 & 1 & 1 \\
\midrule
total & 20 & 13 & 3 & 0 & 4 (20\%) & 5 (25\%) \\
\bottomrule
\end{tabular}
\end{table}

In this agent and under this projection, thirteen of twenty failures are owned by whoever writes the agent's instructions, three by the tool layer, and four, one in five, would be remedied by a different model and nothing else; five under the sensitivity reading. Under either reading the model is the minority owner, for this agent.

\subsection{What the count does not mean}
\label{sec:count-caveats}

\paragraph{Data is never primary, by construction.} The plain-language ``stale filing'' failure exists in these traces: in one replica the agent parsed the FY2024 10-K while the FY2025 10-K sat in the same search result. It lands in policy, because the remedy the analyzer names is an instruction to check the period. The projection sends retrieval failures to data only when the analyzer says the material could not reach the window; when it says the agent did not look, the owner is the policy. Zero in the data column is a property of the projection and of this agent, not evidence that retrieval never failed.

\paragraph{The same question fails differently.} Table~\ref{tab:same-q} shows the primary layer of each test question across the three replicas. PFE is policy three times; CZR is tool, policy, tool; VSEC is model, policy, policy. Part of this is the agent failing in different ways (in one VSEC replica it rounds 60.41 to 60 basis points; in another it divides by segment revenue instead of consolidated), and part is the instrument (\S\ref{sec:errors}).

\begin{table}[t]
\caption{Primary layer of each failed test question across the three replicas (reading A). ``--'' = passed.}
\label{tab:same-q}
\small
\begin{tabular}{@{}llll@{}}
\toprule
question & r1 & r2 & r3 \\
\midrule
ASPI & policy & tool & policy \\
CRM & model & policy & policy \\
CZR & tool & policy & tool \\
FTAI & policy & policy & model \\
MAR/WH & model & policy & -- \\
PFE (disclosure) & policy & policy & policy \\
VSEC & model & policy & policy \\
\bottomrule
\end{tabular}
\end{table}

\section{Can you trust the diagnosis?}
\label{sec:panel}

The count in \S\ref{sec:count} is only as good as the analyzer that produced it. We checked it two ways: blind, against readers that never saw its output, and against itself. Three things are being checked and they come out differently: what failed, where in the trace, and which layer owns the remedy.

\subsection{Blind panel}
\label{sec:blind}

Four blind readers from three model families (Claude Opus 5, Gemini 3.1 Pro, Gemini 3.8 Flash, GPT-5.5) each received the same 20 failed trajectories with the question, the failed criteria and the numbered trace (for the three API readers, a length-capped rendering of it), and nothing from the analyzer. Each named the decisive event and the fault. A separate judge (Claude Opus 5), shown only pairs of justifications, unlabelled and in random order, judged whether the two described the same wrong quantity or the same wrong choice. The protocol, including how ``same fault'' is decided, was written before the panel ran (Appendix~\ref{app:protocols}).

\begin{table}[t]
\caption{Blind panel, 20 failed test trajectories. Same fault: judged on unlabelled justification pairs; under the pre-registered rule, which also counts decisive events within $\pm$5, the two totals are 75/80 and 113/120. Within $\pm$5: decisive event within five trace events.}
\label{tab:panel}
\small
\begin{tabular}{@{}lrr@{}}
\toprule
pair & same fault & within $\pm$5 \\
\midrule
analyzer vs Gemini Pro & 17/20 & 10/20 \\
analyzer vs Gemini Flash & 19/20 & 11/20 \\
analyzer vs GPT-5.5 & 20/20 & 10/20 \\
analyzer vs Opus & 19/20 & 14/20 \\
\midrule
analyzer vs readers & \textbf{75/80} & 45/80 \\
readers vs readers (6 pairs) & \textbf{112/120} & 84/120 \\
\bottomrule
\end{tabular}
\end{table}

\paragraph{What failed: corroborated.} The analyzer names the same fault as a blind reader in 75 of 80 cases; the readers agree among themselves in 112 of 120 (Table~\ref{tab:panel}). Every one of the 20 trajectories has its fault confirmed by at least three of the four readers.

\paragraph{Where: looser on both sides.} The analyzer's decisive event is within five trace events of a reader's in 45 of 80 pairs; the readers' are within five of each other's in 84 of 120. The analyzer is as good as an independent reader at saying what went wrong and about as imprecise at saying where.

\paragraph{Which layer: a definition, not a perception.} Asked to place the fault in our four layers, the blind readers said model in 66 of 80 labels and policy once. In 12 of the 20 Opus labels the reader quotes a violated instruction and still writes model. The readers answered \emph{who erred}; the table answers \emph{who owns the remedy}. We report this as a definition, not as a disagreement: the owner-of-the-remedy reading is ours, and it is not the reading a model gives unprompted. Given the rule, it is reproducible: a second reader with Rule M, the trace and the decisive index, and none of the analyzer's prose, returned the same verdict on all eight cells where the rule applied.

The five misses are informative. Two are Gemini Pro calling ASPI and CZR benchmark artefacts rather than agent failures; on ASPI the reader is wrong (\S\ref{sec:errors}). Two are CRM, where the reader names a different arithmetic error that the analyzer did not report. One is CZR, where the reader blames a wrong-basis choice and the analyzer a corrupted extraction.

\subsection{The analyzer's own errors}
\label{sec:errors}

\paragraph{An over-claim, caught by a sister replica.} On CZR the analyzer wrote that ``correcting this single decision point [\ldots] would flip every downstream EBITDAR/growth/CAGR check''. The sister replica had made that decision and still failed all eight checks. Three replicas took three different bases for the same figure and each was 0.6--2.1\% off, against a 0.5\% tolerance. No figure any replica retrieved reproduces the rubric. A reviewer reading one trajectory would have believed the analyzer; the replica did not.

\paragraph{Same input, two taxonomies.} Given a byte-identical payload twice, the analyzer returned nine categories once and eight the next time, and the two categories outside the prompt layer became one, of a different kind. The layer distribution barely moved, which favours classifying by owner of the remedy over classifying by the analyzer's category. But the instability reaches the count through one field: in one replica the analyzer attached \texttt{verification} to three of its four prompt categories, Rule M applied only to the fourth and returned policy, and the model count was zero; in another it attached none to six of eight, Rule M fired, and the model count was three. The per-replica spread in Table~\ref{tab:count} is that.

\paragraph{Same fault, four labels.} In four of the five analyses of the accuracy arm (\S\ref{sec:acc}) the analyzer named the stale-period failure, each time under a different category label.

\paragraph{A benchmark artefact that was not.} ASPI never passes its must-pass criteria under any configuration. The analyzer emits a data-issue category on it and one blind reader calls it a benchmark artefact, for a visible reason: the question asks for the gross margin between 2/28/2024 and 2/28/2025, and ASP Isotopes reports on calendar years. The periods are those of Renergen, a South African producer that ASP Isotopes agreed in May 2025 to acquire, and whose audited statements for the years ended 28 February 2025 and 29 February 2024 ASP Isotopes filed as an exhibit to an 8-K. Those statements reproduce the rubric: plant depreciation in cost of sales rises from R1.5 million to R34.2 million, +2{,}149.47\%, and the gross margin falls 88.62 points; one normalised figure in the rubric we could not reproduce. In one baseline replica the agent's own EDGAR search returned that exhibit (event 39), and the agent chose ASP Isotopes' quarterly figures instead (event 40). The failure is the agent's, and the trajectory stays in the count, where the frozen rule had already kept it. An analyzer, a blind reader and, for a while, the authors took a hard question for a broken one; the trace said otherwise.

\paragraph{Replay.} We replayed the analyzer twelve times over one scored dev run: two models, with the full run and with one failed question alone, three repetitions each (Appendix~\ref{app:errors}). The stronger model returned six to seven categories where the production model returned three to five; given one question in isolation, both placed the AMZN stale-filing failure in a different \texttt{fix\_level} in different repetitions (prompt, orchestration, architecture, tool description), and one repetition stated as trace evidence an inference the trace does not contain.

What this section supports is bounded. The analyzer is corroborated on what failed. The layer is our definition and is reproducible across model families when the rule is given. The instrument's failure modes are known: it over-claims counterfactuals, it is unstable on the harness-component axis, and it names the same fault inconsistently.

\section{Can you trust a change?}
\label{sec:loop}

We report two runs of the loop as shipped on the 18 dev questions, one under a cost objective and one under an accuracy objective (Fig.~\ref{fig:protocol}). Each round, the analyzer reads the traces of the current dev run; the loop picks one of its findings, proposes one change on one lever and states the trace evidence the change answers; the change is applied, the dev split re-run once, and the gate compares the new run with its parent. Under the cost objective a change is promoted if the bill falls by at least 5\% and the pass rate stays within 10 points of the session's first run; under the accuracy objective, if the pass rate rises by at least one point and the bill rises by no more than 50\%. A session stops after five rounds or two consecutive rejections. The gate reads one run, and \S\ref{sec:drift} shows how far one run moves: it filters changes, it does not prove them. What the loop guarantees is that every change answers something observed in a trace; whether it works is left to the test. The loop's own final configuration, one configuration and no choice at numbers seen, then goes to the frozen test split, three replicas, against three baseline replicas, under a decision rule written before the test was opened (Appendix~\ref{app:protocols}). We read every promoted prompt change before the test for leakage: no company, ticker, figure, date or specific filing may appear. Both winners passed, with one declared exception noted below. The two arms are read together: the same rule, written in advance, and the same measured noise band are applied to both; the cost change clears the band and the accuracy change does not.

\subsection{Cost arm}
\label{sec:cost}

Five rounds (Table~\ref{tab:epochs-cost}). The analyzer's spend reading found the nested retrieval call at 74--96\% of the cost of every question it sampled, and the agent's own reasoning a minor share; the loop moved that call, a document extraction, from Sonnet 4.5 to Haiku 4.5 (round 1, cost 0.437$\times$ its parent, promoted). In traces where the agent queried the same stored document three to eight times, it then tried to teach the agent to batch its sub-questions into one retrieval call per document, first as a tool description (round 2, 0.996$\times$, rejected) and then as a system-prompt paragraph (round 3, 0.844$\times$, promoted). Round 4 tried to chain arithmetic into a single calculator call and saved four percent, rejected. Round 5 answered a trace with ten EDGAR searches, five or six of them near-repeats, for a 10-K that was never filed, before the agent found the answer in an 8-K; it added one sentence, to pivot to an 8-K exhibit or web search after two or three empty EDGAR queries, promoted.

The same instruction saved nothing as a tool description and 16\% of the bill as a system-prompt line. In this agent, the layer, not the wording, owned the remedy.

\begin{table}[t]
\caption{Cost arm, dev split: five rounds, gate = cost ratio against the round's parent.}
\label{tab:epochs-cost}
\scriptsize
\begin{tabular}{@{}llrrl@{}}
\toprule
round & lever & cells & \$ (ratio) & verdict \\
\midrule
base & as shipped & 105 & 31.79 & \\
1 & retrieval model $\to$ Haiku 4.5 & 120 & 13.90 (0.437) & promoted \\
2 & tool desc.: batch per document & 103 & 13.85 (0.996) & rejected \\
3 & system prompt: batch per document & 117 & 11.73 (0.843) & promoted \\
4 & system prompt: chain arithmetic & 117 & 11.24 (0.958) & rejected \\
5 & system prompt: pivot after empty EDGAR & 104 & 10.53 (0.898) & promoted \\
\bottomrule
\end{tabular}
\end{table}

\begin{table}[t]
\caption{Accuracy arm, dev split: four rounds then plateau, gate = cells against the round's parent.}
\label{tab:epochs-acc}
\scriptsize
\begin{tabular}{@{}llrrl@{}}
\toprule
round & lever & cells ($\Delta$) & \$ & verdict \\
\midrule
base & as shipped & 109 & 29.42 & \\
1 & sys.\ prompt: scope qualifiers & 116 (+7) & 33.31 & promoted \\
2 & sys.\ prompt: five ambiguity rules & \textbf{130} (+14) & 38.09 & promoted \\
3 & sys.\ prompt: re-read item by item & 113 ($-$17) & 36.98 & rejected \\
4 & sys.\ prompt: reconcile multi-year & 117 ($-$13) & 36.81 & rejected \\
\bottomrule
\end{tabular}
\end{table}

\subsection{Accuracy arm}
\label{sec:acc}

Four rounds, then a plateau (Table~\ref{tab:epochs-acc}). Every change the loop chose was a system-prompt paragraph. Round 1 added scope qualifiers and gained seven cells, promoted; seven is the distance between baselines 6 and 7 in Table~\ref{tab:drift}, and the gate promoted it. Round 2 answered five dev traces in which the agent had already retrieved the right figures and chose the wrong scope of an ambiguous term at one step (gross against net interest, all cash against cash in excess of the required minimum, equity price against enterprise value); it added five rules for these recurring ambiguities and reached 130 of 149, above every one of the seven shipped baselines. The gains sat where the rules pointed: TGT/WMT went from 3 to 8 cells on the net-interest rule, EPAM from 2 to 5 on the minimum-cash rule. On dev, the intervention landed where the diagnosis had pointed.

Two rounds later the loop stopped. Both rejected rounds kept the round-2 rules and scored TGT/WMT 3 of 8 both times and EPAM 2 and 1 of 5. We wrote, before the test ran, that 130 was one draw of a configuration whose level was unknown. The winning configuration's second paragraph carries one example figure (``79 million shares'') that names no issuer; we declared it and sent the configuration to the test.

\subsection{Both arms on the frozen test}
\label{sec:test}

Table~\ref{tab:test} and Fig.~\ref{fig:main}c--d give the test result for both arms: 9 questions, 3 replicas of the winner against 3 of the baseline, paired bootstrap over questions, the cost ratio judged against 0.83, the widest twin ratio of totals in the dev drift.

\begin{table*}[t]
\caption{Both loop winners on the frozen test split, 9 questions $\times$ 3 replicas, pooled. Intervals are paired bootstraps over the nine questions (10{,}000 resamples). The noise band for cost is [0.83, 1/0.83], the widest dev twin ratio; the accuracy rule is whether the interval excludes zero. Verdicts are those the pre-registered rule returns.}
\label{tab:test}
\footnotesize
\begin{tabular}{@{}lrrrrrrrll@{}}
\toprule
configuration & cells/270 & pass rate & $\Delta$ pts & 95\% CI & must-pass ok/27 & total \$ & cost ratio & 95\% CI & verdict (accuracy / cost) \\
\midrule
baseline (shipped) & 171 & 0.633 & & & 9 & 71.17 & & & \\
cost winner (round 5) & 161 & 0.596 & $-$3.7 & [$-$8.9, +1.1] & 11 & 24.27 & \textbf{0.341} & [0.216, 0.580] & within noise / \textbf{real saving} \\
cost arm, round 3 (secondary) & 168 & 0.622 & $-$1.1 & [$-$7.4, +5.4] & 12 & 28.26 & 0.397 & [0.325, 0.495] & within noise / real saving \\
accuracy winner (round 2) & 175 & 0.648 & +1.5 & [$-$5.3, +7.2] & 11 & 76.98 & 1.082 & [0.844, 1.505] & within noise / within noise \\
\bottomrule
\end{tabular}
\end{table*}

The cost winner spends 34\% of the baseline, with the whole confidence interval below the noise band, at a pass rate within noise of the baseline. The saving is almost entirely on the nested retrieval side: \$62.91 of the baseline's \$71.17 was nested-LLM spend, against \$15.11 of the winner's \$24.27. Per question (Table~\ref{tab:perq}), the disclosure question on PFE fell from \$22.34 to \$3.95 over three replicas; CRM, the financial-modeling question, is the one the winner makes dearer, because the agent spends 129 turns and 114 single-operation calculator calls on it, the defect the rejected round-4 change had targeted. Two questions, CZR and ASPI, are unsolved by every configuration. The cost protocol also declared, before the test, a secondary configuration: round 3, the promoted configuration with the most dev cells. We report it because it was declared. It returns the same verdicts as the winner, a real saving (0.397$\times$) at a pass rate within noise.

\begin{table}[t]
\caption{Cost arm, per test question, summed over three replicas.}
\label{tab:perq}
\small
\begin{tabular}{@{}lrrrr@{}}
\toprule
question & base cells & winner cells & base \$ & winner \$ \\
\midrule
MAR/WH & 17/21 & 17 & 6.63 & 1.60 \\
CZR & 0/24 & 0 & 13.01 & 2.35 \\
HUBS/NOW/TOST & 27/27 & 27 & 3.81 & 1.74 \\
AER/FTAI/WLFC & 22/39 & 20 & 3.81 & 1.73 \\
PFE/SGEN & 27/27 & 27 & 4.70 & 0.74 \\
ASPI & 4/27 & 6 & 9.26 & 5.33 \\
VSEC & 23/27 & 20 & 2.90 & 1.52 \\
PFE (disclosure) & 33/39 & 26 & 22.34 & 3.95 \\
CRM & 18/39 & 18 & 4.70 & 5.30 \\
\bottomrule
\end{tabular}
\end{table}

The accuracy winner gained four cells of 270. Its +14 points on dev became +1.5 on held-out questions, with a confidence interval that straddles zero and a sign-flip test at $p = 0.74$; its cost is within noise too. The two paragraphs written on dev cases did not carry to nine questions about other issuers: VSEC lost cells in all three replicas, CRM gained five, then none, then lost four.

\subsection{Read together}
\label{sec:together}

Same loop, same judge, same split, same rule, one pass and one fail. The structural remedy, a cheaper nested model and an instruction about how to use a tool, survived questions the loop had never seen. The case-specific remedy, two paragraphs about ambiguities in particular dev questions, did not. Without the drift measurement and the written rule we would have reported ``+14 points on dev'', and it would have been true and worthless. The same instrument that passes one change fails the other, and that is what makes the 0.34 worth reporting.

Two boundaries on this reading. The cost result is consistent with the diagnosis having pointed at a real remedy; it does not isolate the diagnosis, because the loop has no control arm (\S\ref{sec:limits}), and the test split is now closed, so none can be added to this study. And cost is the axis on which the instrument has resolution, 0.34 against a band of 0.83; on accuracy the band is the result.

\section{Trust is an audit trail}
\label{sec:trail}

\subsection{Limits, and what we do not claim}
\label{sec:limits}

Our judge scores the benchmark's criteria; the benchmark's own grader would score differently, and no number here is comparable with the Vals leaderboard. The judge is a single model family, and we have not measured how the pass-rate verdicts move under a second judge. We have not run that check; the stored final answers allow the six test reports to be re-scored by another model without re-running the agent or reopening the split. What we can say now is that the cost result does not depend on the judge's verdicts, and that the judge's three votes were unanimous on all but 4 of 1{,}080 test cells, a statement about within-configuration stability and nothing more. Nine test questions and twenty failed trajectories make an anatomy, not an estimate of a population share; the inferential unit is the question, $n=9$. There is no control arm: the loop is the system as shipped, and the gain belongs to it end to end, not to the diagnosis in isolation; we validate the loop, not the causal efficacy of its diagnosis. One agent shows, and does not prove, that the approach transfers. The blind readers are language models of three families, not financial analysts; the owner-of-the-remedy question for a human analyst is open. The accuracy winner was compared against the baseline replicas of the first opening, reused, so the pair does not share a time window. One example figure survived the leakage read of the accuracy winner's prompt. The dev split's analyses were produced under the cost objective and Rule M was not applied to them. Two substrate failures live in the traces and not in any score: the nested retrieval call fails with a prompt-too-long error when one call spans several large filings, and the agent's provider client waits 57 minutes before noticing a dead connection.

\subsection{Related work}
\label{sec:related}

The Vals Finance Agent Benchmark~\cite{bigeard2025fab} supplies our agent, questions and rubric; its paper, which describes the benchmark's first version, reports the best model at 46.8\% and a logarithmic accuracy--cost relation, and grades answers. BigFinanceBench~\cite{wang2026bigfinancebench} grades the visible derivation against point-weighted rubrics and localises failures along the analyst workflow; it does not name the layer that owns the fix. DFAH-Bench~\cite{khatchadourian2026dfah} replays financial decision tasks and finds tool paths diverging (67--69\% agreement) under agreeing decisions (94--95\%); our \S\ref{sec:drift} measures the same instability on a shipped research agent and \S\ref{sec:anatomy} says where, for this agent, it comes from. Khatchadourian and Franco~\cite{khatchadourian2025drift} measure output drift across providers and propose deterministic harnesses with finance-calibrated invariants; our drift is at the level of a multi-step agent's trajectory, where greedy decoding does not remove it. MAST~\cite{cemri2025mast} taxonomises multi-agent failures by symptom and finds that they often stem from system design rather than the model alone; our taxonomy is by owner of the remedy, on a single agent, and reaches a minority-model conclusion for that agent by a different route.

\section{Conclusion}
\label{sec:conclusion}

We asked whether a shipped finance agent can be trusted and answered with measurements rather than a cure. The score is a draw from a distribution 18 criteria wide. When this agent fails, under our taxonomy, the model owns one fix in five and the harness owns the rest. The diagnosis that says so is corroborated blind by readers of three model families on what failed, and its own failure modes are listed. A change to the agent can be trusted only against the measured drift and under a rule written in advance, and when we did that, the same instrument passed a structural change and failed a case-specific one. Trust, on this evidence, is the trail that lets a reader check each of those sentences against a file. A practitioner can build that trail with nothing from us: freeze a split, run the baseline more than once, write the promotion rule against the measured drift, and read the trace before believing the diagnosis.

\bibliographystyle{ACM-Reference-Format}
\bibliography{refs}

\appendix

\section{Taxonomy: declared choices}
\label{app:taxonomy}
Four choices in Table~\ref{tab:projection} have alternatives we did not take. (1) Tool description maps to tool, not policy: both are text, but the description ships with the tool and the tool's owner fixes it. (2) Orchestration with a control component maps to policy: defensible only because this agent is a thin prompt-driven loop with no coded planner; on a coded orchestrator the cell would be tool. (3) Rule M as written, rather than ``model is whatever the declared model lever would remedy'': the simpler rule ties the taxonomy to the action space, and we kept it trace-based. (4) Data issue is outside the anatomy and reported separately, not a fifth bucket.

Two interpretations were declared when Rule M was applied. A \texttt{submit\_final\_result} tool call is the trace's output event (verified on all 27 test trajectories: the output event's content equals the last such call's argument). Reading A, the count, treats any other tool call and any tool result as failing condition (b); reading B, the sensitivity, additionally treats a calculator call whose arguments carry the wrong figure as model-authored.

\section{Anatomy tables}
\label{app:tables}

Table~\ref{tab:verdicts} lists every failed trajectory of the pre-registered set with its decisive event, the analyzer category holding it, the projected layer under reading A and, where it differs, reading B. Event indices are positions in the ATIF trace. Category names (\texttt{EDGE\_CASE}, \texttt{REASONING\_ERROR}, \ldots) are the analyzer's own labels; the layer columns are ours, obtained through Table~\ref{tab:projection} and Rule M.

\begin{table*}[t]
\caption{Per-trajectory verdicts, three test baseline replicas. ``fix $\times$ comp'' is the analyzer's \texttt{fix\_level} $\times$ \texttt{harness\_component} of the primary category.}
\label{tab:verdicts}
\scriptsize
\begin{tabular}{@{}lllllll@{}}
\toprule
replica & question & decisive event & primary category & fix $\times$ comp & layer A & layer B \\
\midrule
r1 & ASPI & assistant \#20 & EDGE\_CASE & prompt $\times$ none & policy & \\
r1 & CZR & tool\_result \#18 & INTEGRATION\_ERROR & orchestration $\times$ observation & tool & \\
r1 & CRM & submit \#143 & INSTRUCTION\_VIOLATION & prompt $\times$ none & model & \\
r1 & FTAI & calculator \#53 & REASONING\_ERROR & prompt $\times$ none & policy & model \\
r1 & PFE & assistant \#49 & KNOWLEDGE\_GAP & prompt $\times$ context & policy & \\
r1 & MAR/WH & submit \#33 & REASONING\_ERROR & prompt $\times$ none & model & \\
r1 & VSEC & submit \#63 & INSTRUCTION\_VIOLATION & prompt $\times$ none & model & \\
\midrule
r2 & CZR & assistant \#22 & REASONING\_ERROR & prompt $\times$ verification & policy & \\
r2 & CRM & calculator \#41 & REASONING\_ERROR & prompt $\times$ verification & policy & \\
r2 & ASPI & edgar\_search \#3 & TOOL\_CALL\_GAP & tool\_description $\times$ action & tool & \\
r2 & FTAI & calculator \#51 & REASONING\_ERROR & prompt $\times$ verification & policy & \\
r2 & MAR/WH & assistant \#6 & REASONING\_ERROR & orchestration $\times$ control & policy & \\
r2 & PFE & tool\_result \#41 & KNOWLEDGE\_GAP & prompt $\times$ none & policy & \\
r2 & VSEC & submit \#101 & INSTRUCTION\_VIOLATION & prompt $\times$ verification & policy & \\
\midrule
r3 & CZR & tool\_result \#44 & INTEGRATION\_ERROR & tool\_description $\times$ context & tool & \\
r3 & ASPI & assistant \#17 & EDGE\_CASE & prompt $\times$ none & policy & \\
r3 & FTAI & assistant \#110 & INSTRUCTION\_VIOLATION & prompt $\times$ none & model & \\
r3 & CRM & calculator \#94 & REASONING\_ERROR & prompt $\times$ verification & policy & \\
r3 & VSEC & tool\_result \#62 & REASONING\_ERROR & prompt $\times$ verification & policy & \\
r3 & PFE & calculator \#42 & REASONING\_ERROR & prompt $\times$ verification & policy & \\
\bottomrule
\end{tabular}
\end{table*}

The blind panel's per-reader layer labels over the same 20 trajectories: Gemini 3.1 Pro model 14, tool 3, benchmark artefact 3; Gemini 3.8 Flash model 17, tool 2, data 1; GPT-5.5 model 17, data 2, tool 1; Claude Opus 5 model 18, data 1, policy 1. Localisation, same decisive event exactly (before the $\pm$5 relaxation): analyzer vs readers 6, 7, 7, 6 of 20; readers vs readers 12, 9, 9, 10, 9, 11 of 20.

\section{The analyzer's errors: replay}
\label{app:errors}

Twelve analyzer calls over the same scored dev run (a shipped-configuration baseline, 18 trajectories): Claude Sonnet 5 (the analyzer used throughout the paper) and Claude Opus 5, each with the full run and with the AMZN trajectory alone, three repetitions of each condition. Categories returned, full run: Sonnet 4, 5, 3; Opus 6, 7, 6. AMZN alone: Sonnet 2, 2, 2; Opus 4, 4, 3. The \texttt{fix\_level} assigned to the AMZN stale-filing failure across the twelve calls: prompt (7 calls), orchestration (1), architecture (3), tool description (2), with some calls emitting two categories for it. Opus returned four category labels that Sonnet 5 never used:
\begin{itemize}\setlength{\itemsep}{0pt}
\item \texttt{INCOMPLETE\_DISCLOSURE\_COVERAGE}
\item \texttt{INCOMPLETE\_SUPPORTING\_DISCLOSURE}
\item \texttt{UNGROUNDED\_PARAPHRASE}
\item \texttt{QUERY\_MISMATCH}
\end{itemize}
One full-run Sonnet call placed AMZN's decisive step on a filing-selection inference that the trace does not record as an event.

\section{Protocols}
\label{app:protocols}

\paragraph{Test protocol, cost arm (written before any test measurement).} 9 questions, 90 criteria. Judge: Gemini 3.8 Flash, temperature 0, 3 votes. Configurations: the shipped baseline; the round-5 configuration (primary: the loop's last promoted configuration, chosen by the loop's own gate); the round-3 configuration (declared secondary: the promoted configuration with the most cells on dev). Three replicas, interleaved: each replica runs every configuration back to back, order reversed on even replicas, so each pair shares a time window. Metrics declared per measurement (cells/90, severity-weighted, must-pass questions, cost split agent/nested, turns, flags), per replica paired ($\Delta$ cells, cost ratio $\Sigma/\Sigma$, flips) and pooled. Decision rule: accuracy \emph{beats} if the paired bootstrap 95\% CI over questions of $\Delta$ pass rate lies above 0, \emph{worse} if below, otherwise \emph{within run-to-run noise}; the sign-flip $p$ is reported beside it and never used to rescue an interval that includes 0. Cost saving \emph{real} if the CI of the cost ratio lies entirely below 0.83, the widest twin ratio of totals on dev; \emph{within noise} inside [0.83, 1/0.83].

\paragraph{Test protocol, accuracy arm (written after the accuracy optimization ended and before any test measurement of its winner).} Same split, judge, metrics and decision rule. One configuration, the loop's final under the accuracy objective (two system-prompt paragraphs), three replicas, against the three baseline replicas of the first opening, reused. Declared as the second and last opening of the test split. Lever read under the leakage rule passed with one example figure declared.

\paragraph{Blind panel protocol (written before the panel ran).} Each reader receives, per trajectory, the question, the system prompt, the frozen judge's verdict on every criterion, and the numbered trace (length-capped for the API readers); nothing from the analyzer; no access to any other file. Output per trajectory: earliest decisive event index and kind; layer in \{policy, tool, data, model, benchmark artefact\} defined as the owner of the remedy; secondary layers; a 2--5 sentence justification citing event indices; whether the needed facts were already in context; the instruction violated, if any; confidence. Readers: Claude Opus 5, Gemini 3.1 Pro (preview), Gemini 3.8 Flash, GPT-5.5. Same-fault agreement between two diagnoses: decisive steps within $\pm$5 events, or a separate judge (Claude Opus 5, not one of the readers' runs), shown only the two justifications, unlabelled and in random order, says they describe the same wrong quantity or choice. Reported: localisation per reader vs analyzer and per pair of readers; same-fault per pair; layer agreement as the secondary, definitional number.

\section{EDGAR backend substitution}
\label{app:edgar}

The shipped agent's \texttt{edgar\_search} tool queries a paid third-party index of EDGAR. We rebound the tool's execution to the SEC's own full-text search endpoint, keeping name, description, parameters and argument validation unchanged. Check: the 53 \texttt{edgar\_search} calls of one baseline run, with the third-party results as recorded, replayed against the SEC endpoint. On the 40 queries non-empty on both backends: URL-set Jaccard mean 0.85, median 1.00; 21 of 40 identical result sets; same first result in 27 of 40. Of the third-party index's main documents, 315 of 350 were returned by the SEC endpoint (90\%); of its exhibits, 502 of 541 (93\%); of the main documents of the target company named in the query, 81 of 81. Fourteen main documents were returned only by the SEC endpoint (amendments, PDF renditions, subsidiary lists). On shared records, accession number, CIK, form type and filing URL are identical; the filing date differs by one business day on 137 records because the third-party index reports the acceptance timestamp and the SEC endpoint the official filing date. One tokenisation difference: the SEC endpoint treats an apostrophe as part of the token. Six of the seven dev baselines and every test measurement ran on the SEC endpoint; the first dev baseline ran on the third-party index.

\end{document}